\documentclass[runningheads]{llncs}

\usepackage{eccv}

\usepackage{eccvabbrv}

\usepackage{graphicx}
\usepackage{booktabs}

\usepackage[accsupp]{axessibility}  

\usepackage{hyperref}

\usepackage{orcidlink}

\begin{document}

\title{OphEdit: Training-Free Text-Guided Editing of Ophthalmic Surgical Video} 

\titlerunning{OphEdit}

\author{Ritul Jangir$^\star$\inst{1} \and Arkya Jyoti Bagchi$^\star$\inst{1} \and 
Mangalton Okram\inst{1}\and
Saurabh Seetaram Korgaonkar\inst{2}\and
Aiman Farooq\inst{1}\orcidlink{0000-0002-3238-7962}\and
Deepak Mishra\inst{1}\orcidlink{0000-0002-4078-9400}} 

\authorrunning{R.~Jangir et al.}
\institute{Indian Institute of Technology Jodhpur, 342030 India \\
\email{farooq1.@iitj.ac.in,dmishra@iitj.ac.in} \and
All India Institute of Medical Sciences Delhi, 110029}
\maketitle

\begingroup\renewcommand\thefootnote{
}

\footnotetext{$^\star$ Contributed equally}
\endgroup
\begin{abstract}
  High-fidelity surgical video generation holds great promise for medical
  training and AI development, yet precise, controllable editing of such
  videos remains a formidable challenge. Modifying surgical attributes such
  as instrument, tissue interactions or procedural phases is particularly
  difficult due to the strict anatomical and temporal constraints of
  ophthalmic procedures. We introduce OphEdit, the first training-free
  framework for text-guided editing of ophthalmic surgical videos. Our
  approach uses a deterministic second-order ODE inversion pipeline to
  capture attention Value tensors from the source video and selectively
  injects them into the conditional branch of classifier-free guidance
  during denoising, preserving the intricate anatomical geometry of the eye
  while mapping text-driven semantic modifications onto the video. Clinical
  evaluation by an ophthalmologist shows that OphEdit handles complex
  transformations, such as instrument swaps and procedural variations, with
  superior structural fidelity and temporal consistency compared to adapted
  general-domain video editors, offering a scalable route to diverse
  surgical training data. The code and prompts can be accessed at https://github.com/ophedit/OphEdit

\end{abstract}

\section{Introduction}
\label{sec:intro}
Ophthalmic surgical videos are an indispensable resource for medical
education, surgical training, and the development of AI-driven assistance
systems. Recordings of procedures such as cataract surgery capture
fine-grained instrument manoeuvres, tissue responses, and the sequential
flow of surgical phases, and recent AI models can already parse them to
understand workflow, assess surgeon performance, and guide robotic
assistants \cite{mezzina2025surgeons}. Training and validating such systems,
however, demands large quantities of annotated video covering diverse
surgical variations, rare complications, and subtle deviations in technique.

Acquiring surgical video data at this scale remains difficult. Strict
patient privacy regulations, ethical constraints, and the high cost of
expert annotation limit both the volume and diversity of available data,
while clinically critical but infrequent events, such as posterior capsule
rupture, are chronically underrepresented. To mitigate this bottleneck,
text-guided video generation has emerged as a promising strategy for
synthesising surgical content from natural language descriptions
\cite{li2025ophora}.

Recent text-to-video (T2V) diffusion models, including CogVideoX
\cite{yang2024cogvideox}, HunyuanVideo \cite{kong2024hunyuanvideo}, and
Open-Sora \cite{zheng2024open}, achieve impressive realism in natural
domains. In the medical domain, Endora \cite{li2024endora} and Bora
\cite{sun2024bora} explore generative models for endoscopy and broader
biomedical video, while Ophora \cite{li2025ophora} adapts CogVideoX to
ophthalmic surgery through large-scale transfer learning, producing
realistic cataract surgery clips from text prompts. These approaches,
however, operate by regenerating entire sequences from scratch. This
regeneration-based paradigm is computationally expensive and offers little
fine-grained control over specific procedural attributes. What clinically
constrained settings need instead is the ability to \emph{edit} an existing
surgical video, \eg, to swap one instrument for another or alter a single
procedural step, while rigorously preserving the delicate anatomical
geometry of the eye. Controllable, text-guided editing of ophthalmic
surgical videos remains largely unexplored.

Training-based editing frameworks typically require paired data or
task-specific fine-tuning, which is impractical in the surgical domain.
Training-free methods such as FateZero \cite{qi2023fatezero}, UniEdit
\cite{bai2025uniedit}, and FLATTEN \cite{cong2023flatten} instead leverage
pre-trained diffusion models, but applying them directly to surgical video
distorts delicate anatomical structures and violates the strict temporal
continuity of instrument motion. To bridge this gap, we propose OphEdit,
whose contributions are threefold:

\begin{itemize}
\item We present the first training-free video editing framework designed
  for ophthalmic surgery, enabling precise, language-driven manipulation of
  procedural elements such as instrument type, tissue appearance, and
  surgical phase.
\item We develop a deterministic second-order ODE inversion scheme that
  extracts attention Value ($V$) tensors from a source video and selectively
  injects them into the conditional branch of classifier-free guidance
  during denoising, preserving anatomical fidelity while executing
  text-prompted edits.
\item We introduce a second-order midpoint integration strategy for both
  inversion and generation. By computing dual-pathway latent updates at the
  log-SNR midpoint, OphEdit eliminates the jittering artifacts of
  first-order methods, ensuring the smooth procedural dynamics essential
  for clinical realism.
\end{itemize}

\section{Related Works}

\subsection{Natural Video Editing}
Video editing through textual data has been progressing in different directions, from zero-shot approaches to tuning a ViT using 2D images. Zero-shot approaches using pretrained diffusion models for editing without retraining or per-video optimization have been achieved by FateZero~\cite{qi2023fatezero}, UniEdit~\cite{bai2025uniedit}, and FLATTEN~\cite{cong2023flatten}.

FateZero~\cite{qi2023fatezero} achieves this by storing self- and cross-attention maps during DDIM inversion, retaining structural and motion information, and reinjecting them during denoising while minimizing semantic leakage. UniEdit~\cite{bai2025uniedit} extends the tuning-free approach beyond appearance to motion editing, introducing separate reconstruction and motion-reference branches that inject spatial and temporal self-attention into the main editing path, so appearance and motion can be controlled largely independently. FLATTEN~\cite{cong2023flatten} improves frame consistency by incorporating optical flow directly into the attention module, constraining patches along the same flow trajectory to attend to each other so the same physical point stays consistent across frames.

SAMA~\cite{zhang2026samafactorizedsemanticanchoring} factorizes instruction-guided video editing into Semantic Anchoring -- jointly predicting semantic tokens and video latents at sparse anchor frames for instruction-aware structural planning -- and Motion Alignment, which pre-trains the same backbone on motion-centric restoration tasks (cube inpainting, speed perturbation, tube shuffle) so temporal dynamics are learned directly from raw video rather than injected externally. EditVerse~\cite{ju2026editverseunifyingimagevideo} unifies image/video generation and editing as a single token-sequence model, using full self-attention for in-context learning across tasks and modalities, which lets it transfer editing knowledge from abundant image data into video. ViFeEdit~\cite{yu2026vifeeditvideofreetunervideo} targets the video-supervision bottleneck by decoupling spatial from temporal token modeling in a DiT, so the spatial-editing parameters can be tuned purely on 2D image pairs while the base model's pretrained temporal attention stays untouched, preserving motion coherence without any video training data.
\subsection{Surgical Video Generation}

Text-guided end to end synthesis methods like Ophora\cite{li2025ophora} and SurGen\cite{cho2024surgentextguideddiffusionmodel} are built on top of already existing T2V  models rather than building surgery- specific architecture. Ophora , built on Ophora160K dataset of video-instruction pairs, specifically is designed for ophthalmic surgical video generation, using progressive video-instruction tuning to transfer spatiotemporal knowledge from natural domain T2V model to ophthalmic surgical domain. SurGen\cite{cho2024surgentextguideddiffusionmodel} adapts CogVideoX\cite{yang2024cogvideox} architecture to laparoscopic cholecystectomy , generating surgical videos with higher resolution and longer duration than prior works which were conditioned only on coarse phase-level text without structural intermediates.

Beyond text conditioning, a second line of work injects structured intermediates — segmentation maps, scene graphs, or symbolic rules — to gain finer control over anatomy and instrument placement than free-form prompts allow. HieraSurg\cite{biagini2025hierasurghierarchyawarediffusionmodel} decomposes generation hierarchically: a first diffusion stage predicts panoptic segmentation maps from surgical phase and action-triplet labels, and a second stage renders video conditioned on these maps, giving explicit control over anatomy and instrument layout that pure text conditioning lacks. SG2VID\cite{sivakumar2025sg2vidscenegraphsenable} pushes controllability further by conditioning on scene graphs instead of segmentation maps, letting a user directly edit entity size, motion, or the entrance of new tools — control that HieraSurg\cite{biagini2025hierasurghierarchyawarediffusionmodel}'s dense maps do not expose as easily.
The most recent of these structured-conditioning approaches, SWoMo\cite{sivakumar2026swomoneurosymbolicworldmodel}, pushes control a step further by separating what happens from what it looks like: a symbolic component — a rule-based simulator paired with scene graphs — governs tool-tissue motion, while a separate diffusion component is responsible only for rendering visual realism. This split keeps motion physically grounded even for tool trajectories that fall outside the training distribution, something purely learned motion models struggle with, marking a shift toward decoupling physical plausibility from visual synthesis in surgical simulation.

Data scarcity itself has also been treated as a first-class problem rather than a side effect. SurV-Gen\cite{venkatesh2025missionbalancegeneratingunderrepresented} addresses this directly with a lightweight two-stage diffusion design that decouples spatial from temporal modeling, allowing it to reuse pretrained image-diffusion weights and generate convincing clips of underrepresented surgical actions or instruments from only a handful of examples — footage that then improves downstream recognition models trained on the augmented data.
Whether conditioned on text, structured priors, or optimized for data efficiency, these methods share a common goal: producing new surgical footage from scratch. None of them address editing an existing surgical video while keeping its underlying anatomy and instrument identity intact — the gap OphEdit fills.

\section{Methodology}

\begin{figure*}[tb]
    \centering
    \includegraphics[width=\textwidth]{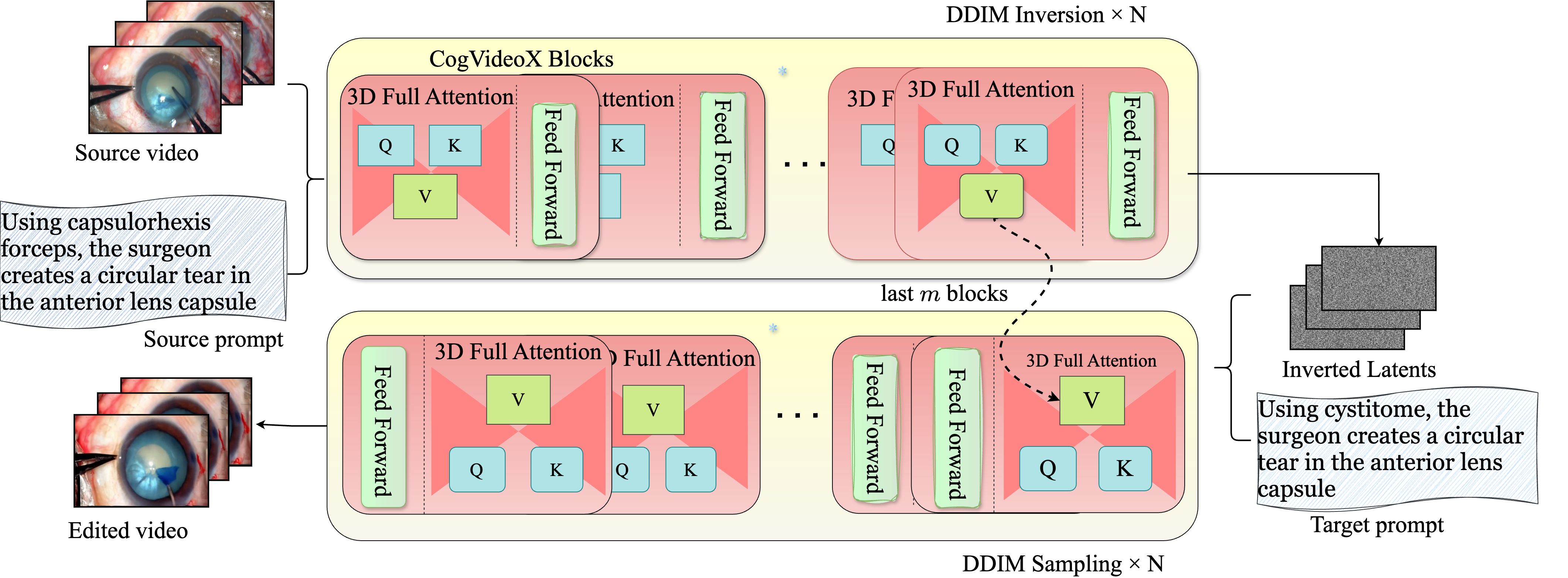}
    \caption{Overview of our proposed training-free video editing
    framework. In the inversion stage, second-order DDIM inversion captures
    the attention Value tensor at both the full-step and midpoint
    evaluations. In the denoising stage, the reverse process is guided by
    the target prompt with timestep-aligned injection of the stored
    values.}
    \label{fig:architecture}
\end{figure*}

OphEdit is a training-free framework for text-guided editing of ophthalmic
surgical videos. Figure~\ref{fig:architecture} shows the overall pipeline:
(1) deterministic inversion of a source video to extract and store
structural attention tensors, and (2) text-guided denoising from the
inverted noise, where the stored tensors are selectively injected to
preserve anatomy while applying the semantic edit.

\subsection{Backbone: CogVideoX and Ophora}
\label{sec:backbone}

OphEdit builds on CogVideoX-2B~\cite{yang2024cogvideox}, a diffusion
transformer for text-to-video generation, and its ophthalmic-domain
variant Ophora~\cite{li2025ophora}. Video frames are compressed by a 3D
causal VAE (8$\times$ spatial, 4$\times$ temporal) into a latent
$z \in \mathbb{R}^{T\times H\times W\times C}$; text prompts are encoded by
a frozen T5-XXL encoder and concatenated with video tokens along the
sequence dimension. CogVideoX employs joint 3D full attention, so every
token attends to every other token across all frames and patches, rather
than factorising attention into separate spatial and temporal blocks; this
is what lets our editing intervention target the attention Value ($V$)
tensors directly, since every frame's content passes through the same
joint attention operation. Ophora adapts this backbone to ophthalmic
surgery via a two-stage fine-tuning process on a large corpus of narrated
cataract-surgery clips: transfer pre-training instils general ophthalmic
domain knowledge (instrument appearances, anatomical structures,
phase-specific dynamics), followed by fine-tuning on a privacy-filtered
subset to align the output distribution with clinical privacy standards.
The resulting model generates high-fidelity, anatomically plausible
surgical videos conditioned on detailed text prompts. OphEdit uses this
frozen Ophora model both to generate source videos and as the backbone for
editing; no parameters are updated at any stage of our pipeline.

\subsection{Deterministic Inversion with Second-Order Integration}
\label{sec:inversion}

Given a source video $V$ generated by Ophora from prompt $p_s$, we encode
it into latent space via the VAE to obtain $z_0$. The goal of inversion is
to recover a noise latent $z_T$ such that denoising from $z_T$ with $p_s$
reconstructs $z_0$. Standard DDIM inversion~\cite{song2020ddim} provides a
deterministic forward process, advancing the latent at each step as
\begin{equation}
    z_{t_{i+1}} = \sqrt{\bar\alpha_{t_{i+1}}}\, \hat z_0
    + \sqrt{1-\bar\alpha_{t_{i+1}}}\, \hat\epsilon_\theta(z_{t_i}, t_i, p_s),
    \label{eq:ddim}
\end{equation}
where $\hat\epsilon_\theta$ is the model's noise prediction, $\bar\alpha_t$
is the cumulative noise schedule, and $\hat z_0$ is the clean latent
estimate derived from $z_{t_i}$ and $\hat\epsilon_\theta$. This is an Euler
discretization of the underlying probability flow ODE, with local
truncation error $O(\Delta t^2)$ that accumulates across the trajectory; in
ophthalmic surgical video this manifests as high-frequency jitter that
violates the continuous physics of instrument motion.

\textbf{Why second-order integration.} Simply increasing the number of
first-order steps only reduces this error linearly, and matching a
second-order solver's $O(\Delta t^3)$ accuracy this way costs roughly twice
the neural function evaluations (NFEs) of running a genuine
predictor-corrector step at the original step count. Adaptive solvers such
as DPM-Solver++~\cite{Lu_2025} reach second-order accuracy
efficiently for sampling, but rely on non-fixed, adaptively chosen
intermediate timesteps; since our editing stage keys stored attention
tensors to specific timesteps and must inject them at those same steps
during denoising, this adaptive scheduling is incompatible with our
requirements. We therefore adopt a fixed-schedule second-order midpoint
scheme (Heun's method), which reaches the same asymptotic accuracy at
equal compute while preserving exact timestep alignment between inversion
and editing.

For each macro-step $t_i \to t_{i+1}$, we perform: (1) a \emph{predictor}
pass that computes the full-step noise prediction
$\hat\epsilon_{\text{full}} = \hat\epsilon_\theta(z_{t_i}, t_i, p_s)$ and
its clean latent estimate $\hat z_0^{\text{full}}$, then takes a temporary
Euler step to the log-SNR midpoint $t_{\text{mid}}$ to obtain an
intermediate state $z_{\text{mid}}$; (2) a \emph{corrector} pass that
evaluates the slope at this intermediate state,
$\hat\epsilon_{\text{mid}} = \hat\epsilon_\theta(z_{\text{mid}}, t_{\text{mid}}, p_s)$
and $\hat z_0^{\text{mid}}$; and (3) an \emph{averaging} step, where
$\hat z_0^{\text{avg}} = (\hat z_0^{\text{full}} + \hat z_0^{\text{mid}})/2$
and
$\hat\epsilon^{\text{avg}} = (\hat\epsilon_{\text{full}} + \hat\epsilon_{\text{mid}})/2$
are substituted into the DDIM update in Eq.~(\ref{eq:ddim}) to reach
$z_{t_{i+1}}$.

The midpoint timestep is chosen in the linearised log-SNR space rather
than by simple index averaging, since the DDPM schedule is non-linear and
$(t_i+t_{i+1})/2$ would not correspond to the physical midpoint of the
noise transition:
\begin{equation}
    \lambda(t) = \tfrac12\log\!\left(\frac{\bar\alpha_t}{1-\bar\alpha_t}\right),
    \qquad
    t_{\text{mid}} = \arg\min_{k}\left|\lambda(k) - \tfrac12\big(\lambda(t_i)+\lambda(t_{i+1})\big)\right|,
    \label{eq:midpoint}
\end{equation}
snapped to the nearest available integer timestep, since the transformer
cannot process fractional timesteps without falling into
out-of-distribution states.

\textbf{Attention storage.} During both the full-step and midpoint
evaluations, we extract the attention Value tensors from the final $m$
transformer blocks, the blocks operating at the highest spatial
resolution, where the model captures fine-grained surgical detail such as
instrument boundaries and delicate tissue textures, and store them in an
attention dictionary keyed jointly by timestep and sub-pass:
$S[t_{i+1}]_{\text{full}}$ for the predictor pass and
$S[t_{i+1}]_{\text{mid}}$ for the corrector pass. Storage is restricted to
the final $n$ inversion steps, the high-noise region in which the global
structural layout of the surgical scene is established. We store only the
Value tensors, not the Query or Key projections: $Q$ and $K$ primarily
govern attention routing, whereas $V$ carries the content, the anatomical
structures and textures of the eye, that we want to preserve.

\subsection{Timestep-Aligned Value Injection for Editing}
\label{sec:injection}

To generate the edited video $V'$, we initialise the reverse diffusion
trajectory from the inverted noise latent $z_T$, conditioned on the target
prompt $p_t$, and run the identical second-order midpoint scheme so that
the inversion and editing trajectories remain symmetric and the stored
keys stay aligned throughout.

\textbf{CFG branch isolation.} Classifier-free guidance (CFG) computes an
unconditional and a conditional noise estimate at each step from two
separate forward passes. This separation gives us a critical degree of
freedom: we can intervene in the conditional branch alone. Let
$V_{\text{stored}}$ denote the attention Value tensor retrieved from
$S[t]$ (the full-step or midpoint key, matching the current sub-pass) at
timestep $t$. We replace the conditional branch's own $V$ tensor with
$V_{\text{stored}}$ in the final $m$ transformer blocks, while its Query
and Key projections, still computed from the target prompt $p_t$, and the
entire unconditional branch are left untouched:
\begin{equation}
    \hat\epsilon_\theta = \hat\epsilon_{\text{uncond}}
    + \omega\big(\hat\epsilon_{\text{cond}}(V_{\text{stored}}) - \hat\epsilon_{\text{uncond}}\big),
    \label{eq:cfg-inject}
\end{equation}
where $\omega$ is the guidance scale. Because $Q$ and $K$ in the
conditional branch continue to be computed from $p_t$, the model still
attends according to the target prompt's semantics; only the content being
attended to ($V$) is anchored to the source anatomy. This has two
consequences. First, \emph{anatomical preservation}: the unconditional
branch, being unmodified, continues to represent the marginal distribution
learned from the source video and anchors the spatial layout and global
anatomy of the eye. Second, \emph{semantic editing}: the conditional
branch is forced to follow $p_t$, but its Value tensors remain constrained
to the source geometry, so the guidance term
$\omega(\hat\epsilon_{\text{cond}} - \hat\epsilon_{\text{uncond}})$ drives
the sample toward the region where both the target prompt and the original
structure are simultaneously satisfied.

\textbf{Injection schedule.} We confine injection to a binary schedule
over the first $n_{\text{inject}}$ denoising steps, the high-noise steps
that govern global structure and spatial composition:
\begin{equation}
    \mathcal{I}(i) = \mathbf{1}_{\{i \le n_{\text{inject}}\}} =
    \begin{cases}
        1, & \text{if } i \le n_{\text{inject}}, \\
        0, & \text{otherwise.}
    \end{cases}
    \label{eq:schedule}
\end{equation}
When $\mathcal{I}(i) = 0$, the model denoises freely under $p_t$, refining
semantic detail and fine texture without further structural constraint.
When $\mathcal{I}(i) = 1$, we intervene in the joint attention layers of
the final $m$ blocks to enforce structural alignment between the source
and target latents.

\textbf{Second-order denoising with injection.} At each timestep $t$, the
injection is executed twice: a full-step update at $(z_t, t)$ using the
full-step lookup key $S[t]_{\text{full}}$ anchors the initial trajectory,
followed by a midpoint update at the intermediate latent $z_{\text{mid}}$
and its log-SNR midpoint $t_{\text{mid}}$, using the midpoint lookup key
$S[t]_{\text{mid}}$ to refine the motion estimate. The final latent
$z_{t-1}$ is derived by averaging the two sub-step predictions, mirroring
the averaging step used during inversion. This symmetrical formulation
maintains a consistent ODE trajectory~\cite{song2020ddim} across inversion
and editing, eliminating the flickering artifacts of first-order
integration and preserving the continuous procedural dynamics essential
for clinical realism. The edited latent $z_0'$ is decoded through the
pre-trained spatio-temporal VAE decoder $\mathcal{D}$
(Section~\ref{sec:backbone}), where $V' = \mathcal{D}(z_0')$ encapsulates
the textually guided edits while retaining the source video's anatomical
and temporal priors.

\section{Experiment and Results}

\subsection{Dataset and Implementation Details}
\label{sec:implementation}
We evaluate OphEdit using the Ophora-28K dataset~\cite{li2025ophora}, a
privacy-preserving subset of the larger Ophora-160K collection. Ophora
provides a small set of 12 canonical prompts, one per procedural
phase, spanning Anterior Chamber Flushing, Capsule Polishing,
Capsulorhexis, Hydrodissection, Incision, Irrigation-Aspiration, Lens
Implantation, Lens Positioning, Phacoemulsification, Tonifying
Antibiotics, Viscoelastic Injection, and Viscoelastic Suction, which are
used to condition the pre-trained Ophora model for source video
generation. We treat each canonical prompt as a source prompt $p_s$ and
generate the corresponding source video with the pre-trained Ophora
text-to-video model; since OphEdit is fully training-free, none of these
prompts or videos are used for fine-tuning.

To construct editing targets, an expert ophthalmologist modified each of
the 12 base prompts to introduce a specific, clinically meaningful
procedural change while preserving the underlying anatomical context, for
example, replacing a phacoemulsification handpiece with an
irrigation-aspiration handpiece, or substituting sodium hyaluronate with
trypan blue dye during viscoelastic injection.

Generated clips have an average duration of 5.54 seconds, are uniformly
sampled to 49 frames at 16 fps, and are resized to $240 \times 352$ before
processing. Our framework is implemented in PyTorch, built on the
CogVideoX and Ophora codebases. We use $N=150$ inference steps, inject stored attention values during the
first $n_{\text{inject}}$ denoising steps (tuned per target prompt;
$n_{\text{inject}}=10$ by default across our main evaluation set), restrict
injection to the final $m=8$ transformer blocks, and use a guidance scale of
$\omega=7.5$. All experiments were run on NVIDIA 80GB
A100 GPUs.

\subsection{Evaluation Protocol}

Text-guided editing of ophthalmic surgical videos has no established benchmark. Therefore, we adopt a clinical evaluation protocol conducted by an experienced ophthalmologist who is blinded to the model identity. For each edited video $V'$, the evaluator watches the clip alongside its target prompt $p_t$ and assesses how well the video depicts the intended surgical step, action, and clinical intent specified in the prompt, rather than merely whether superficial visual details coincide. Each edited video is evaluated according to the following three criteria using a 3-point scale (0 = completely incorrect, 1 = partially correct, 2 = completely correct):

\begin{itemize}
    \item \textbf{Anatomical Consistency}: Assesses whether the delicate ocular structures (e.g., the cornea, iris, and lens capsule) remain anatomically plausible and free from distortions after editing.
    
    \item \textbf{Correct Operation}: Assesses whether the intended semantic edit (e.g., instrument replacement or dye substitution) is faithfully performed according to the target prompt.
    
    \item \textbf{Overall Edit}: Provides a holistic assessment of the edited video by jointly considering anatomical consistency and semantic accuracy.
\end{itemize}

\begin{figure}[t]
\centering
\begin{minipage}{0.48\linewidth}
    \centering
   \captionof{table}{Comparison of correctness and editing performance reported on a 0--1 scale. Best values in each column are shown in {bold}.}
    \label{tab:res1}
    \fontsize{7}{8}\selectfont
    \setlength{\tabcolsep}{2pt}
    \begin{tabular}{@{}lccc@{}}
    \toprule
    \textbf{Model} & \textbf{Anat.} & \textbf{Op.} & \textbf{Edit} \\
    \midrule
    FateZero~\cite{qi2023fatezero} & \textbf{1.00} & 0.29 & 0.63 \\
    FateZero-v2 & 0.95 & 0.34 & 0.53 \\
    UniEdit~\cite{bai2025uniedit} & 0.68 & \textbf{0.55} & 0.63 \\
    OphEdit (DDIM) & \textbf{1.00} & 0.47 & 0.63 \\
    OphEdit (2nd Order) & \textbf{1.00} & \textbf{0.55} & \textbf{0.66} \\
    \bottomrule
    \end{tabular}
\end{minipage}
\hfill
\begin{minipage}{0.48\linewidth}
    \centering
     \includegraphics[width=\linewidth,height=5cm]{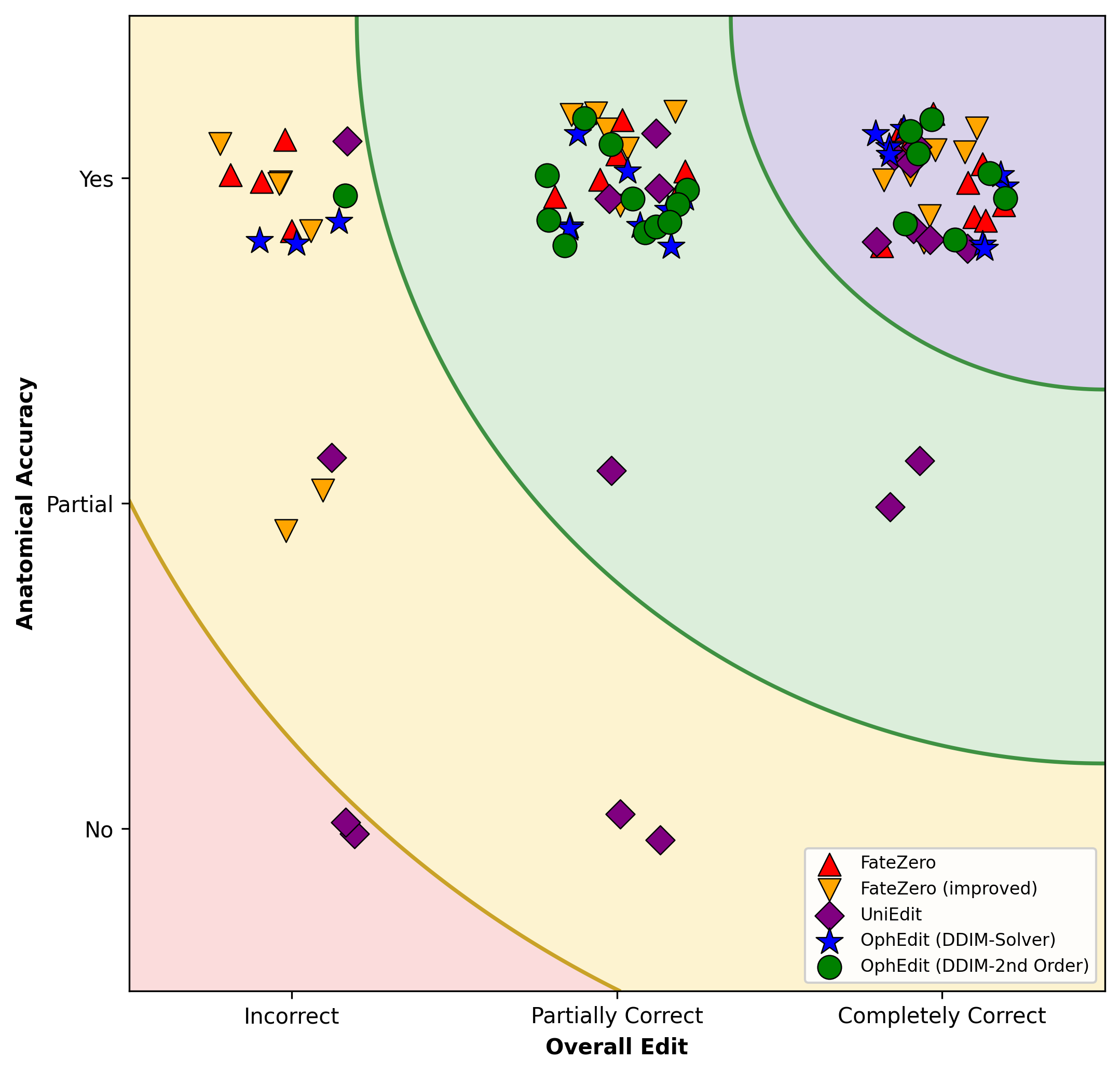}
    \captionof{figure}{Distribution of clinical evaluation scores across phase-specific surgical interventions.}

    \label{fig:shade}
\end{minipage}
\end{figure}
As baselines, we compare against FateZero~\cite{qi2023fatezero} and
UniEdit~\cite{bai2025uniedit}, both state-of-the-art training-free video editing methods originally designed for natural scenes, adapted to the
surgical domain using the same Ophora backbone and the same target
prompts as OphEdit.
FateZero-v2 restricts attention-map injection to the same fixed
final $m$ transformer blocks used by OphEdit, rather than injecting across
all blocks as in the original FateZero, isolating the effect of our
block-selection strategy when applied to FateZero's own attention-based
editing mechanism. We compare two variants of OphEdit: standard first-order
DDIM integration (OphEdit (DDIM)) and our proposed second-order
midpoint scheme (OphEdit (2nd Order)).
\subsection{Results}

Quantitative results are
presented in Table~\ref{tab:res1}. OphEdit (DDIM) and OphEdit (2nd Order)
both achieve the highest Anatomical Consistency (1.00), matched by
FateZero, while UniEdit trails at 0.68 and FateZero-v2 reaches
0.95. On Correct Operation, OphEdit (2nd Order) attains the highest
score (0.55), tied with UniEdit, ahead of OphEdit (DDIM) (0.47),
FateZero-v2 (0.34), and FateZero (0.29). On Overall Edit,
OphEdit (2nd Order) achieves the best score (0.66), narrowly ahead of
FateZero, UniEdit, and OphEdit (DDIM), which are tied at 0.63, with
FateZero-v2 lowest at 0.53. Figure~\ref{fig:shade} visualizes the distribution of clinical evaluation scores across all 19 phase-specific interventions. Each point represents a single edited video; the colored regions demarcate performance zones: purple and green indicate successful edits with preserved anatomy, while yellow and red indicate partial or complete failures. OphEdit (2nd Order) has no points falling in the anatomically
incorrect row, indicating that it never produces a complete
loss of anatomical structure, even on edits that are only partially
correct, demonstrating its ability to perform accurate surgical
modifications while preserving delicate ocular structures.

To complement the clinical evaluation with an automatic, reproducible
metric, we additionally report Fr\'echet Video Distance (FVD)~\cite{unterthiner2018towards} between
each method's edited videos and the distribution of real Ophora source
videos, shown in Table~\ref{tab:fvd_comparison}. OphEdit (DDIM)
achieves the lowest FVD among all edited-video methods (1388), the
closest to the source-video distribution (1192), indicating that its
edits remain the most visually and temporally consistent with real
ophthalmic surgical footage. FateZero\cite{qi2023fatezero} and FateZero-v2 follow
(1483 and 1512, respectively), while UniEdit\cite{bai2025uniedit} records the highest
FVD (1556). Notably, OphEdit (2nd Order) records a higher FVD
(1550) than its first-order counterpart, despite achieving the
strongest clinical scores in Table~\ref{tab:res1}. We attribute this to
the second-order scheme allowing the target prompt to exert a stronger
influence on the denoising trajectory, which improves clinically judged
edit correctness but introduces frame-level appearance shifts that FVD,
which measures distributional similarity to real videos rather than
semantic or anatomical correctness, penalises. This divergence between
FVD and clinical scoring is itself informative: it suggests FVD alone is
an unreliable proxy for editing quality in this domain, and motivates our
reliance on clinical evaluation as the primary measure of success, with
FVD reported as a complementary, reproducible reference point rather than
a substitute.
\begin{table}[h]
\centering
\caption{FVD comparison across models. Best values in each column are shown in {bold}.}
\label{tab:fvd_comparison}
\begin{tabular}{lr}
\toprule
Model  & FVD \\
\midrule
Ophora (source videos) & 1192 \\
FateZero\cite{qi2023fatezero} & 1483 \\
FateZero-v2 & 1512 \\
UniEdit\cite{bai2025uniedit} & 1556 \\
\midrule
OphEdit (DDIM) & \textbf{1388} \\
OphEdit (2nd Order) & 1550 \\
\bottomrule
\end{tabular}
\end{table}
\subsection{Qualitative Results}

\begin{figure*}[]
    \centering
    \includegraphics[width=\textwidth]{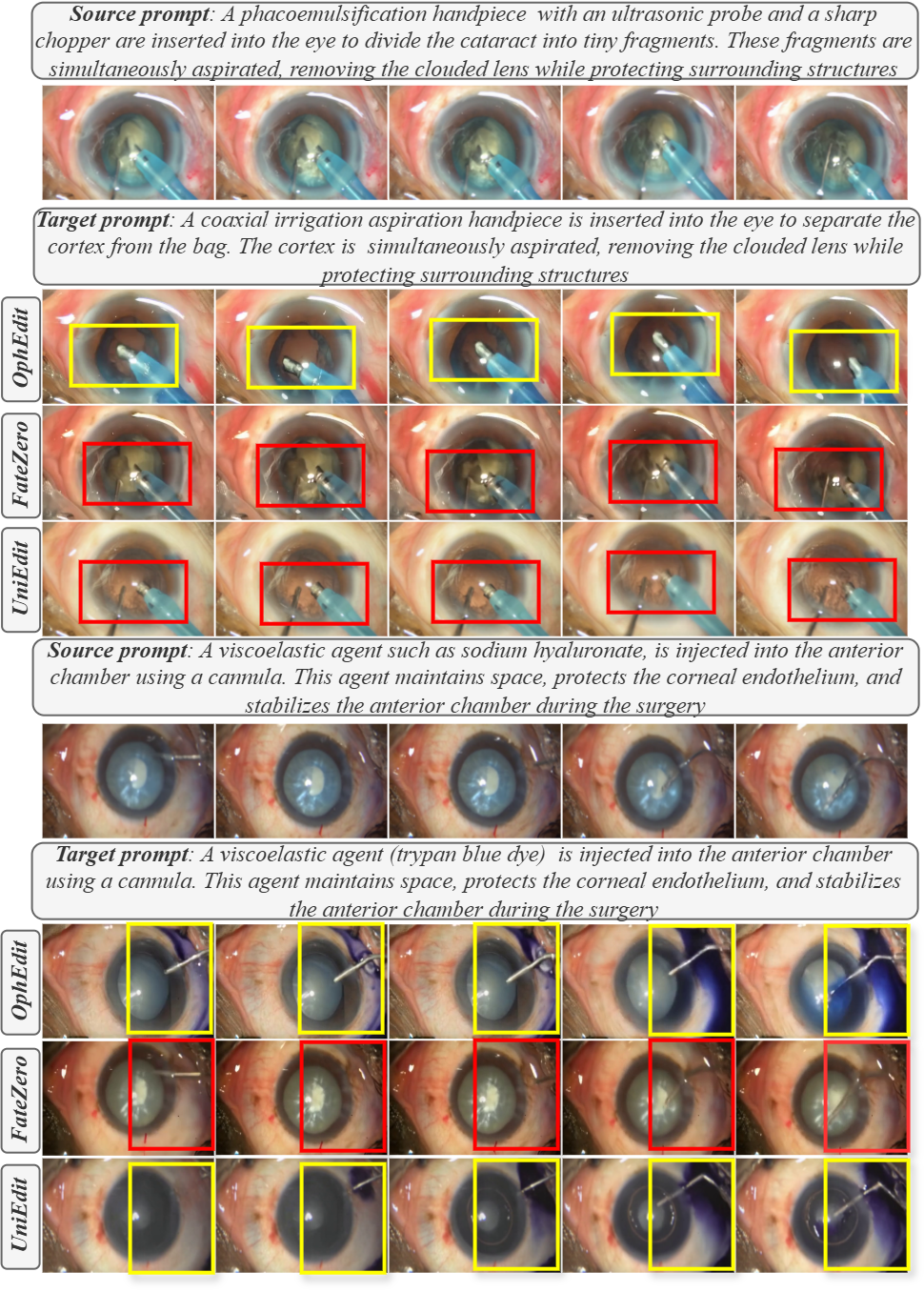}
    \caption{Qualitative comparison of text-guided surgical video editing
    between OphEdit, FateZero and UniEdit . Top: editing from Phacoemulsification
    (source) to Irrigation/Aspiration (target), where the ultrasonic probe
    is replaced with a coaxial irrigation-aspiration handpiece. Bottom:
    editing from Viscoelastic Injection (source) to Trypan Blue Dye
    Injection (target). OphEdit preserves anatomical structures and
    generates clinically plausible instrument transitions, while FateZero
    and UniEdit produces visible artifacts and fails to maintain temporal consistency.
    Yellow boxes indicate correct edits; red indicates incorrect ones.}
    \label{fig:output}
\end{figure*}

Figure~\ref{fig:output} presents a side-by-side comparison of edited video
frames produced by OphEdit, FateZero  and UniEdit for two representative editing
tasks. The top row shows an instrument swap: the source video depicts
phacoemulsification with an ultrasonic probe, and the target prompt
requests replacement with a coaxial irrigation-aspiration handpiece.
OphEdit generates the correct instrument while preserving the surrounding
ocular anatomy; FateZero fails to modify the instrument and introduces
visual artifacts. The bottom row illustrates a dye substitution: the
source video shows injection of clear sodium hyaluronate viscoelastic, and
the target prompt requests trypan blue dye. OphEdit renders the blue tint
of the dye within the anterior chamber while maintaining crisp anatomical
boundaries; FateZero produces inconsistent coloration and blurs the
delicate tissue structures.

\subsection{Ablation Study}

To identify which attention tensor carries the anatomical structure needed for faithful edits, we ablate the choice of tensor(s) injected into the conditional CFG branch during editing (scoped to the last $n{=}14$ timesteps, last $m{=}8$ blocks; all else fixed as in Sec.~\ref{sec:implementation}). Table~\ref{tab:ablation_qkv} shows $K$-only injection barely improves on the uninjected baseline, while $V$-only injection reaches 1.00 Anatomical Consistency and superior Edit quality (0.63 vs.\ 0.11). This confirms $V$ encodes the anatomical content that must be preserved, while $Q$/$K$ mainly govern attention routing — supporting our choice to inject only $V$.

\begin{table}[h]
    \centering
    \caption{Ablation on which attention tensor(s) are injected into the
    conditional CFG branch during editing, evaluated on the same 19
    clinical edit pairs used in Table~\ref{tab:res1}, reported on a 0--1
    scale. Best values in each column are shown in \textbf{bold}.}
    \label{tab:ablation_qkv}
    \fontsize{7}{8}\selectfont
    \setlength{\tabcolsep}{2pt}
    \begin{tabular}{@{}lccc@{}}
    \toprule
    \textbf{Injected tensor(s)} & \textbf{Anat.} & \textbf{Op.} & \textbf{Edit} \\
    \midrule
    None (direct generation)     & 0.11  & \textbf{0.47} & 0.11  \\
    $K$ only                     & 0.13  & \textbf{0.47} & 0.11  \\
    $V$ only (OphEdit(DDIM))     & \textbf{1.00} & \textbf{0.47} & \textbf{0.63} \\
    \bottomrule
    \end{tabular}
\end{table}

\subsection{Discussion}

These qualitative examples, together with the quantitative comparison in
Table~\ref{tab:res1}, point to two design choices behind OphEdit's
behaviour. First, selective Value tensor injection, storing and
re-injecting only the $V$ tensors rather than full attention maps or the
$Q,K$ projections, is intended to balance structural preservation against
editing flexibility; we treat this as a design hypothesis motivated by the
role of $V$ in encoding content rather than attention routing
(Section~\ref{sec:inversion}), rather than as an independently validated
finding. Second, the second-order midpoint integration scheme is designed
to remove the high-frequency jitter that first-order solvers introduce,
which we expect to be reflected in fewer low-similarity outliers and more
temporally coherent instrument motion in the edited videos. 
However, the clinical evaluation also highlighted the inherent challenges; edits requiring complex fluid-tissue interactions, such as Hydrodissection or deep Phacoemulsification, frequently resulted in lower edit quality scores, as modifying the target prompt occasionally disrupted the temporal logic of the surgical sequence. Despite such cases, our method significantly outperformed the baseline by ensuring that even when an operation failed to execute perfectly, the underlying patient anatomy remained largely intact and free of catastrophic generative artifacts, demonstrating its potential for safe and reliable medical data augmentation.

\section{Conclusion}
This paper presents OphEdit, the first fully training-free framework for the text-guided editing of ophthalmic surgical videos. To address the strict anatomical and temporal constraints of the medical domain, we propose a novel inversion-injection pipeline built upon the Ophora foundation model. By capturing Attention Value (V) tensors during deterministic DDIM inversion and selectively injecting them into the conditional Classifier-Free Guidance (CFG) branch, our approach rigorously preserves delicate ocular structures while accurately executing text-driven semantic modifications. Comprehensive evaluations by an ophthalmologist demonstrate that OphEdit successfully performs complex surgical transformations, such as instrument substitutions and phase transitions, with superior structural fidelity compared to adapted general-domain baselines.


%
%
\bibliographystyle{splncs04}
\bibliography{main}

\begin{thebibliography}{10}
\providecommand{\url}[1]{\texttt{#1}}
\providecommand{\urlprefix}{URL }
\providecommand{\doi}[1]{https://doi.org/#1}

\bibitem{bai2025uniedit}
Bai, J., He, T., Wang, Y., Guo, J., Hu, H., Liu, Z., Bian, J.: Uniedit: A
  unified tuning-free framework for video motion and appearance editing. In:
  Proceedings of the 33rd ACM International Conference on Multimedia. pp.
  10171--10180 (2025)

\bibitem{biagini2025hierasurghierarchyawarediffusionmodel}
Biagini, D., Navab, N., Farshad, A.: Hierasurg: Hierarchy-aware diffusion model
  for surgical video generation (2025), \url{https://arxiv.org/abs/2506.21287}

\bibitem{cho2024surgentextguideddiffusionmodel}
Cho, J., Schmidgall, S., Zakka, C., Mathur, M., Kaur, D., Shad, R., Hiesinger,
  W.: Surgen: Text-guided diffusion model for surgical video generation (2024),
  \url{https://arxiv.org/abs/2408.14028}

\bibitem{cong2023flatten}
Cong, Y., Xu, M., Simon, C., Chen, S., Ren, J., Xie, Y., Perez-Rua, J.M.,
  Rosenhahn, B., Xiang, T., He, S.: Flatten: optical flow-guided attention for
  consistent text-to-video editing. arXiv preprint arXiv:2310.05922  (2023)

\bibitem{ju2026editverseunifyingimagevideo}
Ju, X., Wang, T., Zhou, Y., Zhang, H., Liu, Q., Zhao, N., Zhang, Z., Li, Y.,
  Cai, Y., Liu, S., Pakhomov, D., Lin, Z., Kim, S.Y., Xu, Q.: Editverse:
  Unifying image and video editing and generation with in-context learning
  (2026), \url{https://arxiv.org/abs/2509.20360}

\bibitem{kong2024hunyuanvideo}
Kong, W., Tian, Q., Zhang, Z., Min, R., Dai, Z., Zhou, J., Xiong, J., Li, X.,
  Wu, B., Zhang, J., et~al.: Hunyuanvideo: A systematic framework for large
  video generative models. arXiv preprint arXiv:2412.03603  (2024)

\bibitem{li2024endora}
Li, C., Liu, H., Liu, Y., Feng, B.Y., Li, W., Liu, X., Chen, Z., Shao, J.,
  Yuan, Y.: Endora: Video generation models as endoscopy simulators. In:
  International conference on medical image computing and computer-assisted
  intervention. pp. 230--240. Springer (2024)

\bibitem{li2025ophora}
Li, W., Hu, M., Wang, G., Liu, L., Zhou, K., Ning, J., Guo, X., Ge, Z., Gu, L.,
  He, J.: Ophora: a large-scale data-driven text-guided ophthalmic surgical
  video generation model. In: International Conference on Medical Image
  Computing and Computer-Assisted Intervention. pp. 425--435. Springer (2025)

\bibitem{Lu_2025}
Lu, C., Zhou, Y., Bao, F., Chen, J., Li, C., Zhu, J.: Dpm-solver++: Fast solver
  for guided sampling of diffusion probabilistic models. Machine Intelligence
  Research  \textbf{22}(4),  730–751 (2025). \doi{10.1007/s11633-025-1562-4},
  \url{http://dx.doi.org/10.1007/s11633-025-1562-4}

\bibitem{mezzina2025surgeons}
Mezzina, M., De~Backer, P., Vercauteren, T., Blaschko, M., Mottrie, A.,
  Tuytelaars, T.: Surgeons versus computer vision: a comparative analysis on
  surgical phase recognition capabilities: M. mezzina et al. International
  Journal of Computer Assisted Radiology and Surgery  \textbf{20}(6),
  1283--1291 (2025)

\bibitem{qi2023fatezero}
Qi, C., Cun, X., Zhang, Y., Lei, C., Wang, X., Shan, Y., Chen, Q.: Fatezero:
  Fusing attentions for zero-shot text-based video editing. In: Proceedings of
  the IEEE/CVF International Conference on Computer Vision. pp. 15932--15942
  (2023)

\bibitem{sivakumar2025sg2vidscenegraphsenable}
Sivakumar, S.K., Frisch, Y., Ghazaei, G., Mukhopadhyay, A.: Sg2vid: Scene
  graphs enable fine-grained control for video synthesis (2025),
  \url{https://arxiv.org/abs/2506.03082}

\bibitem{sivakumar2026swomoneurosymbolicworldmodel}
Sivakumar, S.K., Johnson, A., Dhingra, A., Frisch, Y., Ghazaei, G.,
  Mukhopadhyay, A.: Swomo: Neuro-symbolic world model for cataract surgery
  simulation (2026), \url{https://arxiv.org/abs/2605.16530}

\bibitem{song2020ddim}
Song, J., Meng, C., Ermon, S.: Denoising diffusion implicit models. arXiv
  preprint arXiv:2010.02502  (2020)

\bibitem{sun2024bora}
Sun, W., You, X., Zheng, R., Yuan, Z., Li, X., He, L., Li, Q., Sun, L.: Bora:
  biomedical generalist video generation model. arXiv preprint arXiv:2407.08944
   (2024)

\bibitem{unterthiner2018towards}
Unterthiner, T., Van~Steenkiste, S., Kurach, K., Marinier, R., Michalski, M.,
  Gelly, S.: Towards accurate generative models of video: A new metric \&
  challenges. arXiv preprint arXiv:1812.01717  (2018)

\bibitem{venkatesh2025missionbalancegeneratingunderrepresented}
Venkatesh, D.K., Funke, I., Pfeiffer, M., Kolbinger, F., Schmeiser, H.M.,
  Weitz, J., Distler, M., Speidel, S.: Mission balance: Generating
  under-represented class samples using video diffusion models (2025),
  \url{https://arxiv.org/abs/2505.09858}

\bibitem{yang2024cogvideox}
Yang, Z., Teng, J., Zheng, W., Ding, M., Huang, S., Xu, J., Yang, Y., Hong, W.,
  Zhang, X., Feng, G., et~al.: Cogvideox: Text-to-video diffusion models with
  an expert transformer. arXiv preprint arXiv:2408.06072  (2024)

\bibitem{yu2026vifeeditvideofreetunervideo}
Yu, R., Tan, Z., Chen, Z., Liu, S., Wang, X.: Vifeedit: A video-free tuner of
  your video diffusion transformer (2026),
  \url{https://arxiv.org/abs/2603.15478}

\bibitem{zhang2026samafactorizedsemanticanchoring}
Zhang, X., Dong, W., Song, Y., Fang, B., Zhang, Q., Wang, J., Chen, F., Zhang,
  H., Feng, H., Lu, Y., Zhou, H., Yuan, C., Wang, J.: Sama: Factorized semantic
  anchoring and motion alignment for instruction-guided video editing (2026),
  \url{https://arxiv.org/abs/2603.19228}

\bibitem{zheng2024open}
Zheng, Z., Peng, X., Yang, T., Shen, C., Li, S., Liu, H., Zhou, Y., Li, T.,
  You, Y.: Open-sora: Democratizing efficient video production for all. arXiv
  preprint arXiv:2412.20404  (2024)

\end{thebibliography}
\end{document}